\documentclass[10pt,twocolumn,letterpaper]{article}

\usepackage{cvpr}
\usepackage{times}
\usepackage{epsfig}
\usepackage{graphicx}
\usepackage{amsmath}
\usepackage{amssymb}

\usepackage{algpseudocode}
\usepackage{algorithm}
\usepackage{subfigure}
\usepackage[pagebackref=true,breaklinks=true,letterpaper=true,colorlinks,bookmarks=false]{hyperref}

\usepackage{multicol}        % used for the two-column index
\usepackage{array}
\usepackage{longtable}
\usepackage[tableposition=below]{caption}
\usepackage{multirow}  
\usepackage{booktabs}
\usepackage{threeparttable}
\cvprfinalcopy % *** Uncomment this line for the final submission

\def\cvprPaperID{0000} % *** Enter the CVPR Paper ID here

\ifcvprfinal\fi

\begin{document}

\title{ChaLearn Looking at People: Inpainting and Denoising challenges\thanks{This is a preprint of an article published in Inpainting and Denoising Challenges. The Springer Series on Challenges in Machine Learning. Springer, Cham. The final authenticated version is available online at: https://doi.org/10.1007/978-3-030-25614-2\_2}}
\author{\parbox{16cm}{\centering
{Sergio Escalera$^1$, Marti Soler$^2$, Stephane Ayache$^3$, Umut Guclu$^4$, Jun Wan$^5$, Meysam Madadi$^6$, Xavier Bar\'o$^7$, Hugo Jair Escalante$^8$, Isabelle Guyon$^9$}\\
{
$^1$ Universitat de Barcelona and Computer Vision Center, Barcelona, Spain \\
$^2$ Universitat de Barcelona, Barcelona, Spain \\
$^3$ Aix Marseille Univ, CNRS, LIF, Marseille, France \\
$^4$ Radboud University, Donders Institute for Brain, Cognition and Behaviour, Nijmegen, Netherlands \\
$^5$ National Laboratory of Pattern Recognition (NLPR), Institute of Automation, Chinese Academy of Sciences (CASIA) \\
$^6$ Compter Vision Center, Spain \\
$^7$ Universitat Oberta de Catalunya, Spain \\
$^8$ INAOE, Mexico \\
$^9$ Universit\'e Paris-Saclay, France, ChaLearn, Berkeley, CA, USA }}
}
\maketitle

\begin{abstract}

Dealing with incomplete information is a well studied problem in the context of machine learning and computational intelligence. However, in the context of computer vision, the problem has only been studied in specific scenarios (e.g., certain types of occlusions in specific types of images), although it is common to have incomplete information in visual data. This chapter describes the design of an academic competition focusing on inpainting of images and video sequences that was part of the competition program of WCCI2018 and had a satellite event collocated with ECCV2018. The \emph{ChaLearn Looking at People Inpainting Challenge} aimed at advancing the state of the art on visual inpainting by promoting the development of methods for recovering missing and occluded information from images and video. Three tracks were proposed in which visual inpainting might be helpful but still challenging: \emph{human body pose estimation}, \emph{text overlays removal} and \emph{fingerprint denoising}. This chapter describes the design of the challenge, which includes the release of three novel datasets, and the description of evaluation metrics, baselines and evaluation protocol. The results of the challenge are analyzed and discussed in detail and conclusions derived from this event are outlined. 
\end{abstract}

\section{Introduction} 
\label{sec:introduction}

The problem of dealing with missing  or incomplete information in machine learning and computer vision arises naturally in many applications. Recent strategies make use of generative models to complete missing or corrupted data. Advances in computer vision using deep generative models have found applications in image/video processing, such as denoising \cite{jain2009natural}, restoration \cite{xu2014deep}, super-resolution \cite{dong2016image}, or inpainting \cite{xie2012image,newson2014video}. 

We focus on image and video inpainting tasks, that might benefit from novel methods such as Generative Adversarial Networks (GANs) \cite{pathakCVPR16context} or Residual connections \cite{he2016deep,mao2016image}. Solutions to the inpainting problem may be useful in a wide variety of computer vision tasks. In this study, we focus on three important applications of image and video inpainting: human pose estimation, video de-captioning and fingerprint recognition. Regarding the former task, it is challenging to perform human pose recognition in images containing occlusions. It  still  is a present problem in most realistic scenarios. Since tracking human pose is a prerequisite for human behavior analysis in many applications, replacing occluded parts may help the whole processing chain. Regarding the second task, in the context of news media, video entertainment and broadcasting programs from various languages (such as news, series or documentaries), there are frequently text captions or embedded commercials or subtitles. These reduce visual attention and occlude parts of frames, potentially decreasing the performance of automatic understanding systems. Finally, %it is a well known problem 
the problem of  validation of fingerprints in images with occlusions which has applications mostly in forensics and security systems. Despite recent advances in machine learning, it is still challenging to aim at fast (real time) and accurate automatic removal of occlusions (text, objects or stain) in images and video sequences. 

In this chapter we study recent advances in each given task. Specifically, we conduct the \emph{ChaLearn Looking at People Inpainting Challenge} that comprised three tracks, one  for each task. Such event had a satellite event collocated with ECCV2018. Herein we explain the challenge setups, provided datasets, baselines and challenge results for each task. The main contributions of this challenge are: the organization of an academic competition that attracted a number of participants to deal with the inpainting problem; the release of three novel datasets that will foster research in visual inpainting; an evaluation protocol that will remain open so that new methodologies can be compared with those proposed in the challenge. The data can be downloaded from:
\begin{itemize}
\item Still images of humans: \\https://chalearnlap.cvc.uab.cat/dataset/30/description/
\item Video decaptioning: \\https://chalearnlap.cvc.uab.cat/dataset/31/description/
\item Fingerprint reconstruction and denoising: \\https://chalearnlap.cvc.uab.cat/dataset/32/description/
\end{itemize}

The remainder of the chapter is organized as follows. In Section~\ref{sec:track1} we explain the  inpainting of  human images track where images  are occluded with random blocks. In Section~\ref{sec:track2} we elaborate on the problem of inpainting of video clips subtitles with different size and color. Next, we describe the inpainting of noisy fingerprints track in Section~\ref{sec:track3}, where  noise is added artificially with the aid  of different patterns. Finally, we discuss conclusions and future directions of research in Section~\ref{sec:conclusion}.

\section{Inpainting still images of humans} \label{sec:track1}

Human subjects are important targets in many images and photographs. In this sense, having an image with clear details and without noise is critical for some applications in uncontrolled environments. For instance, it is quite likely to remove an object that occludes part of the body and replace it with a realistic background. This has an instant application in media production or security cameras. It can also be used as an intermediate task to help other applications like human pose or gesture estimation.

The human pose has many degrees of freedom, giving the human apperance very high variability. This coupled with cluttered backgrounds, different viewpoints and differences in foreground/background illumination, pose great challenge to the task of inpainting parts of the human body in still images.
%Inpainting parts of human body in still images is a challenging task due to cluttered background and high variability in human appearance because of high degrees of freedom in human pose, different viewpoints and probably different illumination in foreground/background. 
The lack of temporal context in highly masked images adds an extra difficulty for these approaches. Therefore, an algorithm must be aware of global context while generating consistent and realistic local patches for masked areas.

In this track we tackle inpainting still images of humans with different level of missing areas. As a standard approach in the inpainting domain, we artificially mask out several blocks around body joints with random position and size. Given an RGB image and its mask in training time, a participant's method must be able to accurately reconstruct the masked region in test time. Evaluation is done based on standard similarity metrics. Given inpainting as an intermediate task, we also evaluate accuracy of each reconstructed image in human pose. That is, a pretrained method is used to estimate body pose based on reconstructed image and an error is computed based on estimated 2D joints.

\subsection{Data}
\begin{figure}[h]
\centering
	\includegraphics[width=0.35\textwidth]{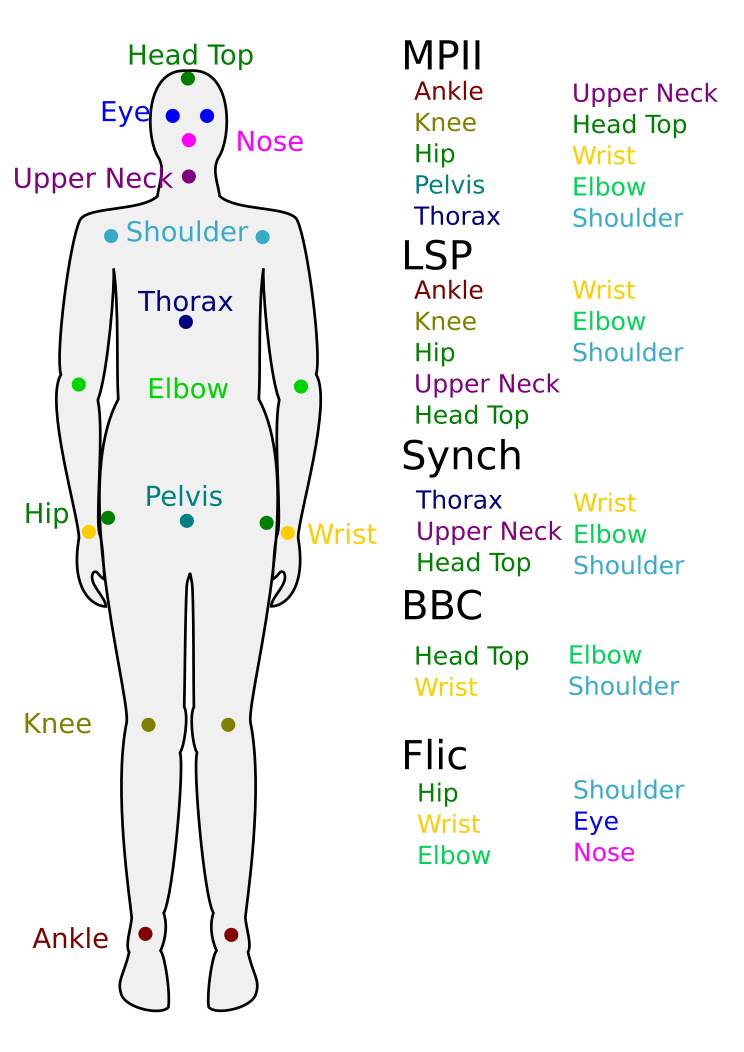}
	\caption{Position of the selected joints.}
	\label{fig:joints}
\end{figure}

For this dataset we collected images from multiple sources that can be seen in Table~\ref{tab:sources}. They have been selected because they provide a high variability of content while being representative enough to allow the resulting methods to generalize. We randomly split samples into train/val/test sets by 70/15/15 \% ratio, respectively.
\begin{table}[h]
		\centering		
                \small
		\caption{Image source overview, note that the number of images used is not the same as the original size of the dataset, this is due to extracting multiple human crops from the original image and filtering out some not suitable for the planned task.}\label{tab:sources} 
		\resizebox{\linewidth}{!}{\begin{tabular}{|l|l|l|}\toprule
			\textbf{Name} & \textbf{\#Images Used} & \textbf{Cropped} \\
			\midrule
			MPII Human Pose Dataset \cite{andriluka14cvpr} & 26571 & Yes \\
			Leeds Sports Pose Dataset \cite{Johnson10}& 2000 & No\\
			Synchronic Activities Stickmen V \cite{eichner2012human} & 1128 & Yes\\
			Short BBC Pose \cite{charles2013domain}& 996 & No\\
			Frames Labelled In Cinema \cite{modec13}& 10381 & Yes\\ 
			\bottomrule
		\end{tabular}}
\end{table} 
\vspace{0.5cm}
\paragraph{Preprocessing}
 To make the joint annotations consistent we made sure that all the selected datasets used the same criteria when annotating each body part. It is important to note that while not every image has all joints annotated, the annotations of all the joints of the same type are consistent. In Figure~\ref{fig:joints} one can see which joint positions we consider valid. In Table~\ref{tab:jointComparision} we can see which joints we extracted from each dataset. The only ones that appear through all the datasets are \textbf{wrists}, \textbf{elbow} and \textbf{shoulder}.
 
%\tiny
\begin{table*}[h!]
		\centering		
		\caption{Check-marks indicate that either the joint appears in some images of the dataset, or it can be extracted from the original annotations. Joints refer to 1-\textbf{Ankles}, 2-\textbf{Knees}, 3-\textbf{Hips}, 4-\textbf{Pelvis}, 5- \textbf{Thorax}, 6-\textbf{Upper Neck}, 7-\textbf{Head Top}, 8-\textbf{Wrists},  9-\textbf{Elbows}, 10-\textbf{Shoulders}, 11-\textbf{Eyes}, 12-\textbf{Nose}}
		\begin{tabular}{|l|l|l|l|l|l|l|l|l|l|l|l|l|l|}\toprule
			\textbf{Name} & \textbf{\#Joints} & \textbf{1} & \textbf{2} & \textbf{3}& \textbf{4} & \textbf{5} & \textbf{6} & \textbf{7}& \textbf{8}& \textbf{9} & \textbf{10} & \textbf{11} & \textbf{12}
			\\
			\midrule
			MPII & 19 & \checkmark & \checkmark & \checkmark & \checkmark &\checkmark & \checkmark & \checkmark & \checkmark & \checkmark & \checkmark& & \\
			LSP & 8 & \checkmark & \checkmark & \checkmark& & &\checkmark & \checkmark& \checkmark & \checkmark & \checkmark & & \\
			Synch & 6 & & & & & \checkmark & \checkmark& \checkmark&\checkmark &\checkmark & \checkmark& &\\
			BBC & 4 & & & & & & &\checkmark &\checkmark & \checkmark& \checkmark& &\\
			FLIC & 6 & & & \checkmark& & & & & \checkmark& \checkmark&\checkmark &\checkmark &\checkmark\\ 
			\bottomrule
		\end{tabular}\label{tab:jointComparision}
\end{table*}

%\normalsize

\noindent
Some adjustments had to be done to ensure that the annotations were consistent:
	
\begin{itemize}
	\item We removed the information about right and left of the joints since some of the datasets were using the camera as reference and others the human subject.
	\item In the Synchronic dataset we did not have a thorax joint, but, from what we could see, it contained a vertical line through the upper body and the middle of that line was pretty similar to the thorax position, so we decided to use those instead.
\end{itemize}
%\vspace{0.5cm}

\paragraph{Masking} 
For each image we generated a different mask to hide parts of the image. Algorithms will be evaluated on how well they can restore the parts of the image occluded by this mask. To avoid having fixed bounds we set the following procedure: masks consist of $N$ blocks, where $0<N<11$ for each image, each block being a square of size $s$ ranging from $$\frac{min(w,h)}{20}<s<\frac{min(w,h)}{3}$$ where $w$ and $h$ are the width and height of the image, respectively. The blocks cover joints in part and are randomly positioned around the center of the joint. The blocks do not overlap and have a margin of $100px$ from the image's edge. At most $70\%$ of the image is masked. 

\subsection{Baselines}
The baselines decided for track 1 are adaptations of ~\cite{pathakCVPR16context, Yang_2017_CVPR, yeh2017semantic} that all involve deep convolutional architectures. Below are descriptions of the considered baselines:
\begin{itemize}
	\item {\bf Context encoders}.~\cite{pathakCVPR16context}. This method uses an autoencoder model along with a generative adversarial model (GAN). It was trained by a conjunction of a reconstruction loss (normalized masked L2) and an adversarial loss provided by the discriminator. 
	\item {\bf Multi-scale neural patches}~\cite{Yang_2017_CVPR}. This baseline is focused on obtaining high resolution results. To do so it uses a model based on two networks, a content network tasked with creating an initial approximation and a texture network which is tasked with adding high frequency details by constraining the texture of the generated image according to the texture of the non masked area. Context Encoders are used as the global content prediction network which serves as initialization for the multi-scale algorithm.
    \item {\bf Semantic inpainting}~\cite{yeh2017semantic}. This model learns to generate new samples from the dataset, which means it learns the manifold where the dataset exists. Then they infer a reconstruction by finding the point on this manifold closest to the encoded masked image.\\
\end{itemize}
We trained these baselines on the same train set used for the competition. Some qualitative results are shown in Table \ref{tab:samples_t1}. As one can see, the baselines perform quite well on small parts but visual quality must be improved when inferring bigger missing regions.

\begin{table*}[htb]
  \begin{center}
  \begin{tabular}{cc@{\hspace{2em}}ccc@{\hspace{2em}}c}
  Image & Masked & \cite{pathakCVPR16context} & \cite{yeh2017semantic} & \cite{Yang_2017_CVPR} & Pose Estimation\\
\includegraphics[width=0.1\textwidth]{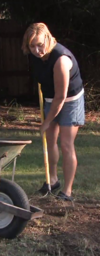} & 
  \includegraphics[width=0.1\textwidth]{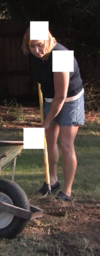} & \includegraphics[width=0.1\textwidth]{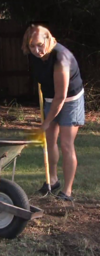} & \includegraphics[width=0.1\textwidth]{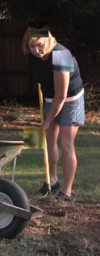} &
  \includegraphics[width=0.1\textwidth]{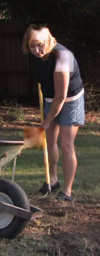} &
  \includegraphics[width=0.1\textwidth]{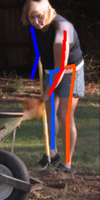}\\
   DSSIM: & - & 0.306 & 0.351 & 0.312 & -\\
  \includegraphics[width=0.1\textwidth]{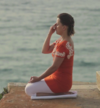} & 
  \includegraphics[width=0.1\textwidth]{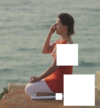} & \includegraphics[width=0.1\textwidth]{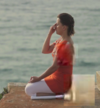} & \includegraphics[width=0.1\textwidth]{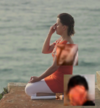} &
  \includegraphics[width=0.1\textwidth]{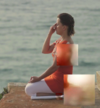} &
  \includegraphics[width=0.1\textwidth]{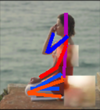}\\ 
  DSSIM: & - & 0.135 & 0.231 & 0.172 & -\\ 

  \includegraphics[width=0.1\textwidth]{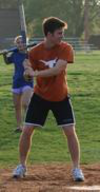} & 
  \includegraphics[width=0.1\textwidth]{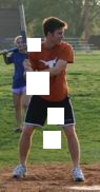} & \includegraphics[width=0.1\textwidth]{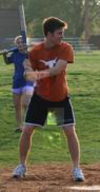} & \includegraphics[width=0.1\textwidth]{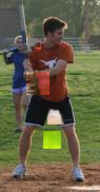} &
  \includegraphics[width=0.1\textwidth]{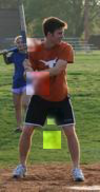} &
  \includegraphics[width=0.1\textwidth]{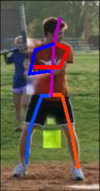} \\
 DSSIM: & - & 0.353 & 0.477 & 0.353 & -\\
  \end{tabular}
  \caption{Qualitative and quantitative samples of chosen baselines for track 1 (resized). DSSIM to the original has been added below the reconstructions. Pose estimation (last column) done over \cite{Yang_2017_CVPR} reconstruction. }\label{tab:samples_t1}
  \end{center}
\end{table*}

\subsection{Competition results}
To evaluate results, we used traditional image similarity metrics such as mean squared error (MSE) and structural dissimilarity (DSSIM)~\cite{wang2004image}. Additionally, we proposed a new highly domain specific metric, Weakly Normalized Joint Distance (henceforth WNJD), which evaluates the performance of a state of the art pose estimation technique on the image before and after the reconstruction. It can be calculated for each image with:
$$WNJD = \frac{\sum_{i=0}^{N}{\left|OJoint_i - PJoint_i\right|}}{N\left|\left|(w,h)\right|\right|},$$
where $OJoint$ and $PJoint$ are vectors containing the original and predicted joints, respectively. $N$ is the number of labeled joints and $w,h$ are the width and height of the image. Finally, due to the emphasis on the human pose domain of the problem the submissions were evaluated using a mean rank of multiple metrics. 

We used \cite{newell2016stacked} as pose estimation algorithm, which presents an architecture called hourglass which is stated to be able to capture information from all possible scales. Each hourglass module is an autoencoder with residual connections from encoder convolutional layers to corresponding decoder ones. A number of hourglass modules are stacked sequentially, a loss is applied on the output of each stack and the whole model is trained jointly. Finally, the joints are extracted from output joint heatmaps. In the competition, this information was withheld from the participants to avoid unfair advantages.

The final results on the competition test set can be seen in Table \ref{tab:testT1Res}. We also show some qualitative images of participants methods in Figure~\ref{tab:qualitative_t1}.

\begin{table}
\centering
\caption{Test set results for track 1.}
\label{tab:testT1Res}
\resizebox{\linewidth}{!}{\begin{tabular}{@{}llllll@{}}
\toprule
\textbf{\#} & \textbf{User} & \textbf{Rank} & \textbf{MSE} & \textbf{DSSIM} & \textbf{WNJD} \\ \midrule
1           & UNLU         & 1.3          & 0.0158 (1)   & 0.2088 (2)   & 0.1489 (1)    \\
2           & Inception       & 1.7           & 0.0158 (2)   & 0.2048 (1)   & 0.1495 (2)    \\
\bottomrule
\end{tabular}}
\end{table}

\begin{table*}[!htbp]
  \begin{center}
  \begin{tabular}{ccc}
  Ground truth & UNLU & Inception\\
\includegraphics[width=0.24\textwidth]{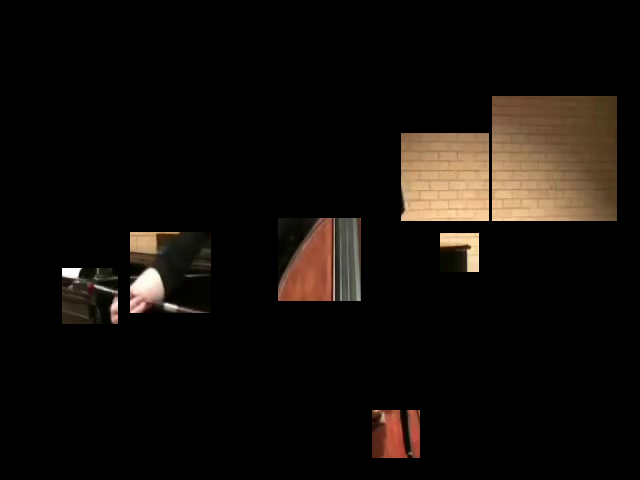} & \includegraphics[width=0.24\textwidth]{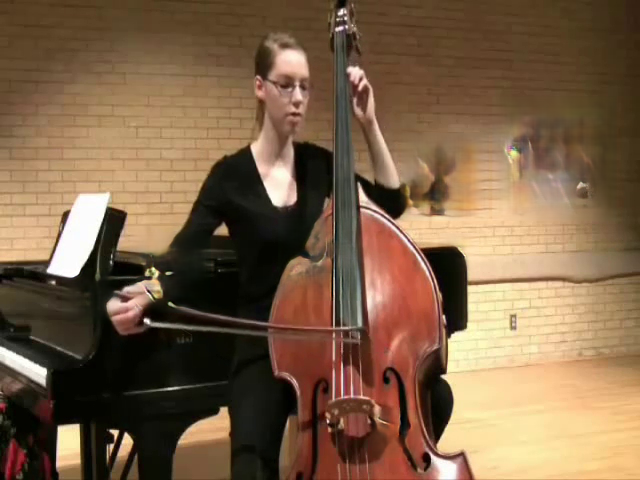} & \includegraphics[width=0.24\textwidth]{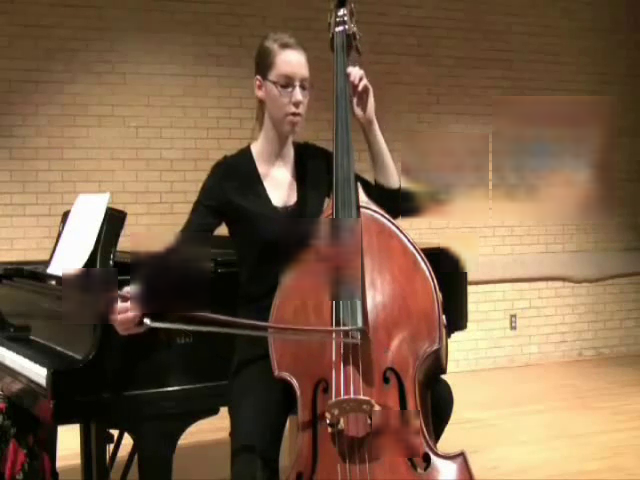} \\
\includegraphics[width=0.24\textwidth]{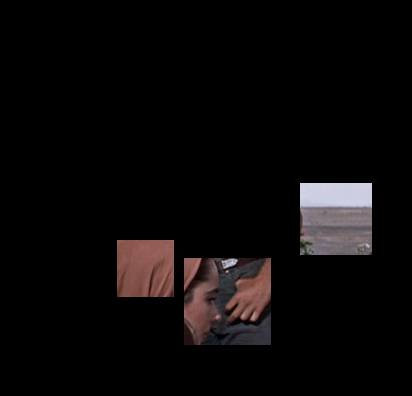} & \includegraphics[width=0.24\textwidth]{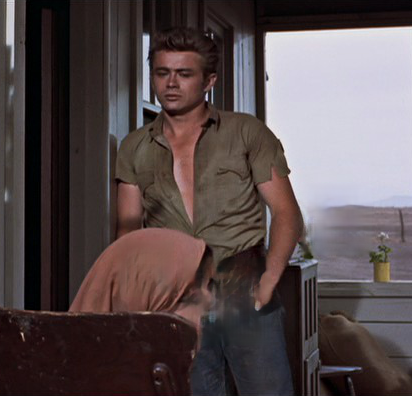} & \includegraphics[width=0.24\textwidth]{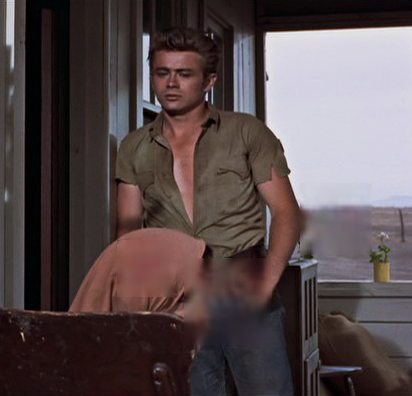} \\
\includegraphics[width=0.24\textwidth]{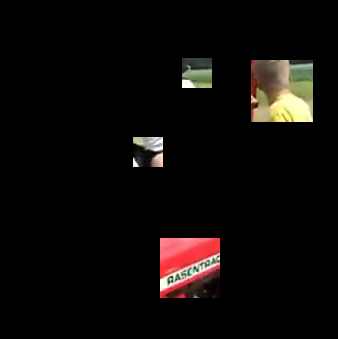} & \includegraphics[width=0.24\textwidth]{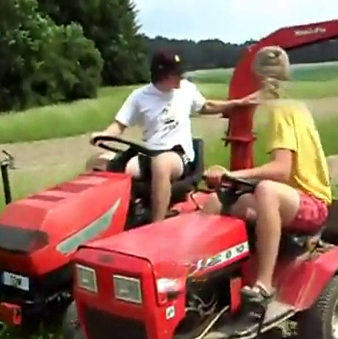} & \includegraphics[width=0.24\textwidth]{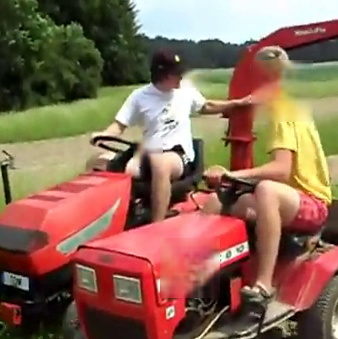} \\
\includegraphics[width=0.24\textwidth]{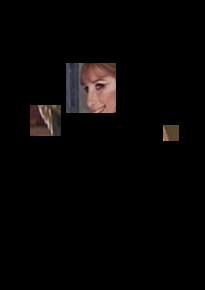} & \includegraphics[width=0.24\textwidth]{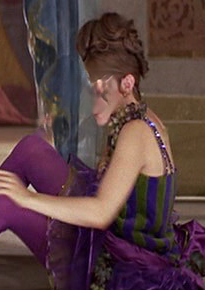} & \includegraphics[width=0.24\textwidth]{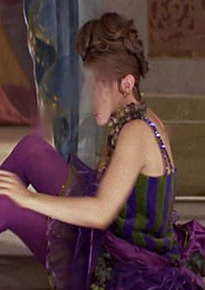} \\
\includegraphics[width=0.24\textwidth]{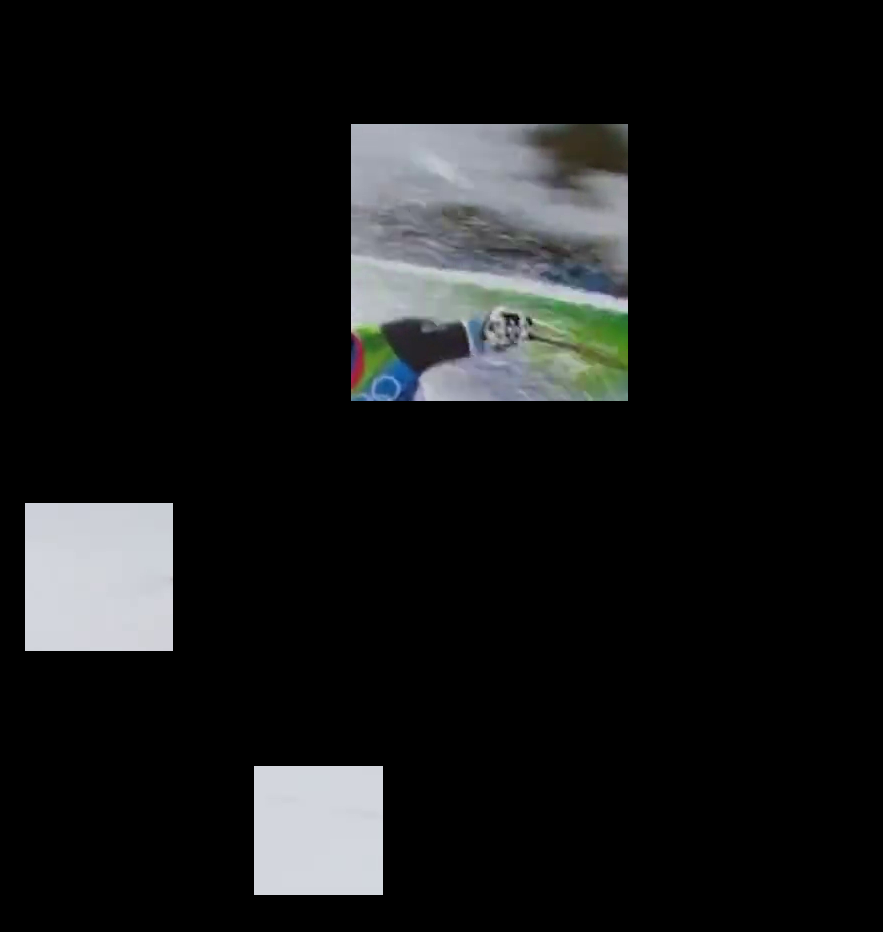} & \includegraphics[width=0.24\textwidth]{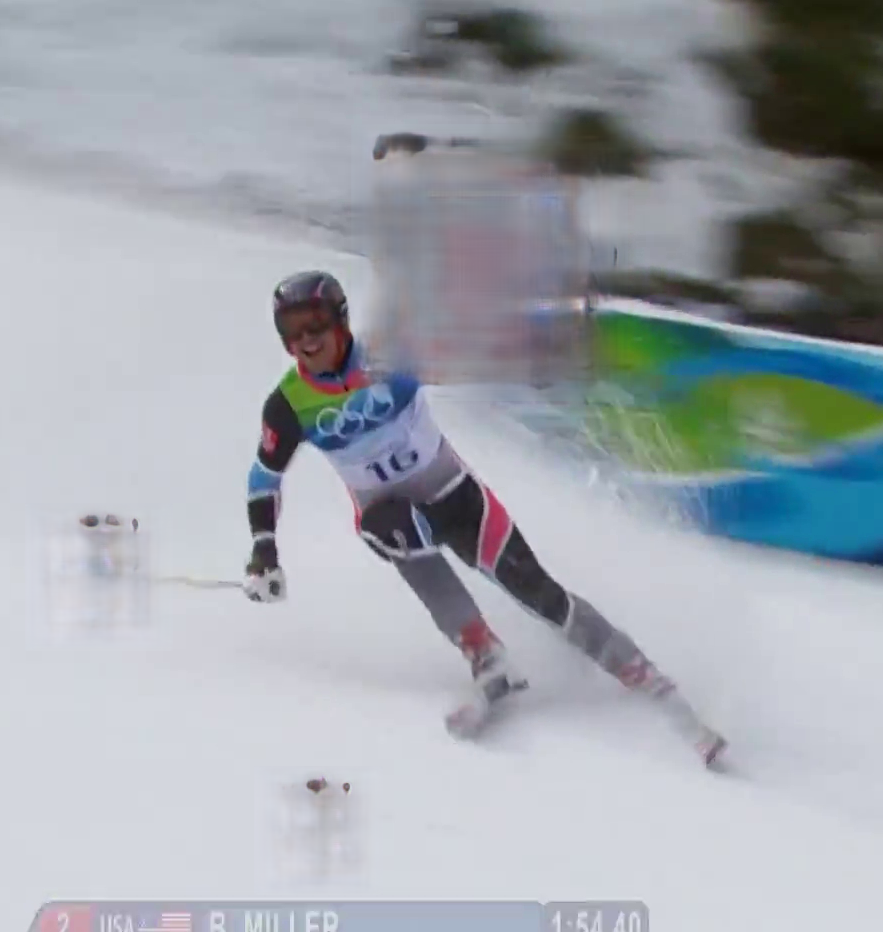} & \includegraphics[width=0.24\textwidth]{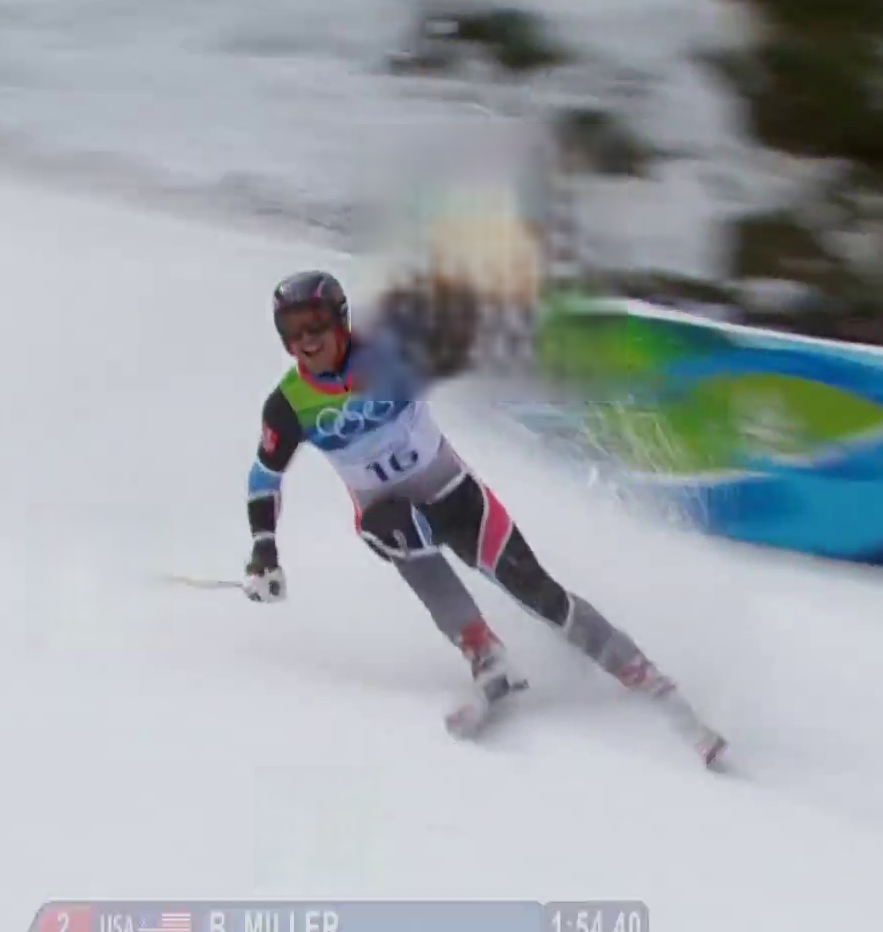} \\
  \end{tabular}
  \caption{Qualitative samples for track 1.}\label{tab:qualitative_t1}
  \end{center}
\end{table*}

\subsection{Discussion}

From the results of the competition we can see that most of the approaches, from participants or baselines, involved the use of a GANs. The main differences with the solutions were on the Loss or mixture of Losses used to train the GAN and the architecture. These differences can be seen in Table~\ref{tab:T1methSumm}.

\begin{table*}
\centering
\caption{Main approach differences between participant approaches.}
\begin{threeparttable}
\resizebox{\textwidth}{!}{\begin{tabular}{@{}lllllcccc@{}}
\toprule
\multirow{2}{*}{\# } & \multirow{2}{*}{Work Name} & \multicolumn{1}{c}{\multirow{2}{*}{Input Size}} & \multicolumn{1}{c}{\multirow{2}{*}{Pretraining}}                                                          & \multicolumn{3}{c}{Generator}                                                                                                                    & \multicolumn{1}{c}{\multirow{2}{*}{Discriminators}} \\ \cmidrule(lr){5-7}
                                                       & \multicolumn{1}{c}{}                           & \multicolumn{1}{c}{}                            & \multicolumn{1}{c}{}                                                                                      & \multicolumn{1}{c}{S\footnotemark[1]} & \multicolumn{1}{c}{D\footnotemark[2]} & \multicolumn{1}{c}{Loss}                                                                      & \multicolumn{1}{c}{}                                
\\ \midrule

1 & \begin{tabular}[c]{@{}l@{}} People Inpainting with \\Generative Adversarial \\ Networks \end{tabular}
& 256x256
& \begin{tabular}[c]{@{}l@{}}Generative Image Inpainting \\with Contextual Attention\\ (Places2)\end{tabular}
& N & Y
& \begin{tabular}[c]{@{}l@{}}Generalized \\ Loss-Sensitive GAN \end{tabular}
& Global + Local \\
\\
2 &\begin{tabular}[c]{@{}l@{}} Generative Image \\Inpainting for Person\\Pose Generation \end{tabular}
& 128x128
& \begin{tabular}{@{}l@{}}None\\ (VGG used to\\ calculate loss)\end{tabular}                               & Y & Y
& \begin{tabular}[c]{@{}l@{}}Mixture:\\ {[}L1 \& Adversarial \\ \& Perceptual(VGG){]}\end{tabular} 
& Global\\
\bottomrule
\end{tabular}}
\begin{tablenotes}
      \footnotesize
      \item[1] Skip-connections
      \item[2] Dilated Convolutions
\end{tablenotes}
\end{threeparttable}
\label{tab:T1methSumm}
\end{table*}

Another interesting point to note is that while both approaches transferred knowledge from networks trained in other domains, they used different methods to do so. One of them used pretraining and the other one incorporated the knowledge into the loss function, by using VGG to evaluate the quality of the reconstruction.\\
We believe that these results indicate that GANs have consolidated their place as a strong competitor for inpainting problems, and that new methods live and die by their ability to correctly evaluate the quality of the reconstruction. We can see the field moving from using the original image as the only solution and instead evaluating each solution on their own merits, independently of how much they resemble the original. This is probably one of the main contributing factors behind the predominance of GANs. We can also see that there is no generalized standard to evaluate these possible solutions which causes most of the methods to focus heavily on loss engineering.\\

In this competition though we did not want to only evaluate the image reconstruction quality. We also wanted to check how capable an inpainting model was to store information from an specific domain such as human poses. To that end we can see a more fine grained breakdown of the mean WNJD for each joint type on Table~\ref{tab:T1joint}

\begin{table*}[]
\centering
\caption{Mean WNJD for all predictions of each \textbf{predicted} joint type in the test set. Lower is better. \textbf{RK} - Right Knee, \textbf{RH} - Right Hip, \textbf{RW} - Right Wrist,\textbf{RE} - Right Elbow, \textbf{RS} - Right Shoulder, \textbf{RA} - Right Ankle, \textbf{P} - Pelvis, \textbf{LE} - Left Elbow, \textbf{LS} - Left Shoulder, \textbf{LK} - Left Knee,\textbf{LH} - Left Hip, \textbf{LW}  Left Wrist, \textbf{T} - Torso, \textbf{LA} - Left Ankle}
\begin{tabular}{@{}llllllllllllllll@{}}
\toprule
\textbf{\#} & \textbf{User} & \textbf{RK} & \textbf{RH} & \textbf{RW} & \textbf{RE} & \textbf{RS} & \textbf{RA} & \textbf{P} & \textbf{LE} & \textbf{LS} & \textbf{LK} & \textbf{LH} & \textbf{LW} & \textbf{T} & \textbf{LA} \\ \midrule
1 & UNLU &  \textbf{0.07} &  \textbf{0.16} &  \textbf{0.24} &  \textbf{0.13} &  \textbf{0.12} &  \textbf{0.16} & \textbf{0.06} &  \textbf{0.11} &  \textbf{0.10} &  \textbf{0.06} &  \textbf{0.09} &  \textbf{0.19} & \textbf{0.08} &  \textbf{0.10} \\
2 & Inception & 0.51 & 0.46 & 0.40 & 0.38 & 0.39 & 0.57 & 0.32 & 0.28 & 0.26 & 0.34 & 0.31 & 0.31 & 0.47 &  0.35  \\
\bottomrule
\end{tabular}
\label{tab:T1joint}
\end{table*}
As we can see there was a definitive difference between the quality of the human pose reconstruction between the approaches, as one model outperformed the other in every type of joint reconstructed. It is notable that no model added any kind of handcrafted feature or used pretrained networks specific to the human domain. All the human domain knowledge was learned from this dataset.

\section{Video decaptioning} \label{sec:track2}
Video data is nowadays collected everywhere, as well as shared on various internet platforms in an extraordinary exponential growing., This offers a wide diversity of contents to final users, e.g. more than 400 hours of new video are uploaded to Youtube every single minute\cite{statista}. Live streaming from social media platforms also offer a huge quantity of video to final users, that for some poor network reasons might be degraded to low resolution. In various situations, available video data might benefit from recent advances in computer vision to enhance its quality and improving the user experience. \\

Existing similar tasks, such as automatic video denoising or video coloration allow for restoring ancient archives that constitute an effective way of preserving such rich source of cultural knowledge. Video inpainting has multiple forms because of the spatio-temporal nature of such signal: in the temporal axis, inpainting might consists in inferring entire frames within a given video sequence, leading to artificial slow motion effect \cite{slomo-nvidia}. For both spatial and temporal axis, video inpainting replaces missing or occluding parts in video sequences with semantically relevant pixels. In particular, removing occluding parts in video sequences might lead to better automatic object detection and indexing systems, especially for autonomous vehicle vision systems. From the user point of view, text overlays in news or media programs reduce the visual attention, making hard to focus on the original video message. \\

This section introduces the ChaLearn LaP Inpainting Challenge Track on video decaptioning. The video decaptioning task might be viewed as a simplified but challenging and realistic scenario of the video inpainting task. We define the video decaptioning task in a supervised setting, where the goal is to generate a decaptioned version of a captioned video sequence. Addressing this task must resolve the complex task of replacing overlaid captions with original set of semantically coherent pixels in the video sequence. We propose the first large dataset of paired training data that contains both degraded and groundtruth data from natural, wide and diverse video sequences. We hope that such large video dataset can push deep learning based model to resolve this challenging task.

\subsection{Data}
\begin{figure*}[htbp] 
\begin{center}
\includegraphics[scale=0.4]{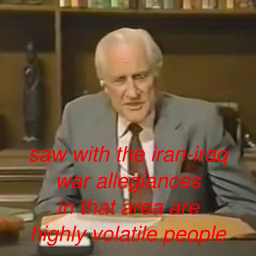}
\includegraphics[scale=0.4]{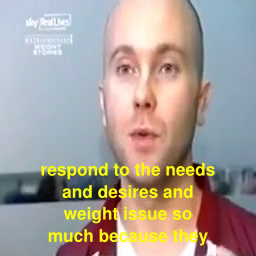}
\includegraphics[scale=0.4]{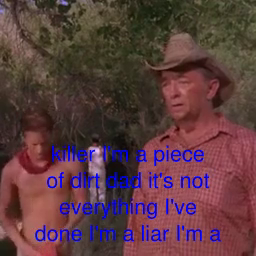}
\includegraphics[scale=0.4]{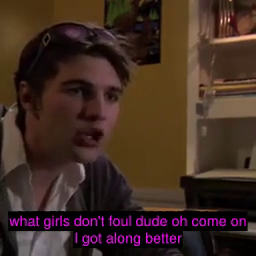}\\
\vspace{0.1cm}
\includegraphics[scale=0.4]{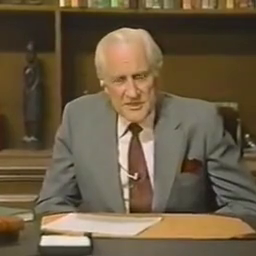}
\includegraphics[scale=0.4]{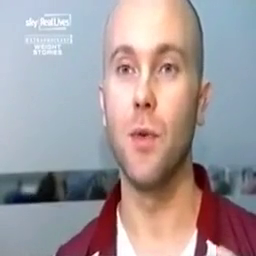}
\includegraphics[scale=0.4]{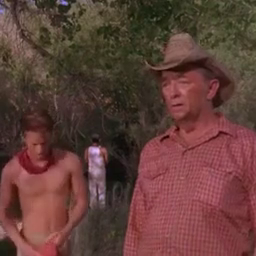}
\includegraphics[scale=0.4]{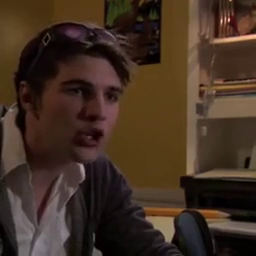}
\caption{Few dataset samples. On top, corrupted X frames. At bottom, groundtruth Y frames}
\label{sample}
\end{center}
\end{figure*}

We collected a large brand new dataset for the video inpainting track of the ChaLearn LaP Inpainting Challenge. The dataset is composed of pairs of the form (X, Y), where X is a 5 seconds video clip (containing embedded text) and Y the corresponding target video clip (without embedded text, see Figure~\ref{sample}). The participants in this track must develop a method that receives as input the video with captions and provide as output the decaptioned video. 

\paragraph{Sources}
We collected about 150 hours of diverse videos of TV movies of various genres with associated generated subtitles available from YouTube. The various video contents are intended to be as diverse as possible, making the dataset almost generic for subtitle removal task. Data sources were checked to fulfill the required copyrights to be reused during this challenge for research purposes.

\paragraph{Preprocessing}
Raw videos were  processed in order to obtain a set of well formed short video sequences: we removed letter-boxing area (top/down black bars where subtitles could be printed), normalized frame rate to 25 frames per second and rescaled frames to width=256 keeping aspect ratio, then cropped the central region with size $256\times256$. In order to create the set X of corrupted videos, we hard printed subtitles by applying several variations in fonts and text style (size, color, italic/bold, background box transparent/opaque). Finally, we split the original and corrupted videos in 5 seconds non-overlapped video clips. 
%\noindent
We kept 70\% of the clips that contain the most subtitles that represents about 90,000 video clips with 125 frames of 256x256 pixels. Finally, we split the resulting video clips into training (60\% of clips), validation (20\% ) and, test (20\%) sets. 

In Figure~\ref{sample} a few examples are shown. The figure illustrates the diversity of video content, style and resolution, as well as the vaeried font transformations and subtitle overlays. We also notice a few examples containing text or logo in both corrupted and ground truth clips.

\subsection{\bf Novelty of the dataset}

This dataset is the largest and more diverse video dataset especially designed for the inpainting task. The nature and the diversity of video content make it both, challenging and of practical relevance,  %in video inpainting, and really valuable to apply in many video domains, as well as for other 
%not only for video processing, but for other fields. %or machine learning tasks, such as scene indexing or video/speech alignment. 
Large variations of text overlays offers various kind of inpainting contributions, from denoising methods (for samples where overlay is small, font is well contrasted and background transparent), to large missing region (where background is opaque and overlay is large). Dealing with such variations and video contents lead to a difficult and challenging dataset. \\

\begin{figure}[h]
\begin{center}
	\includegraphics[width=0.9\linewidth]{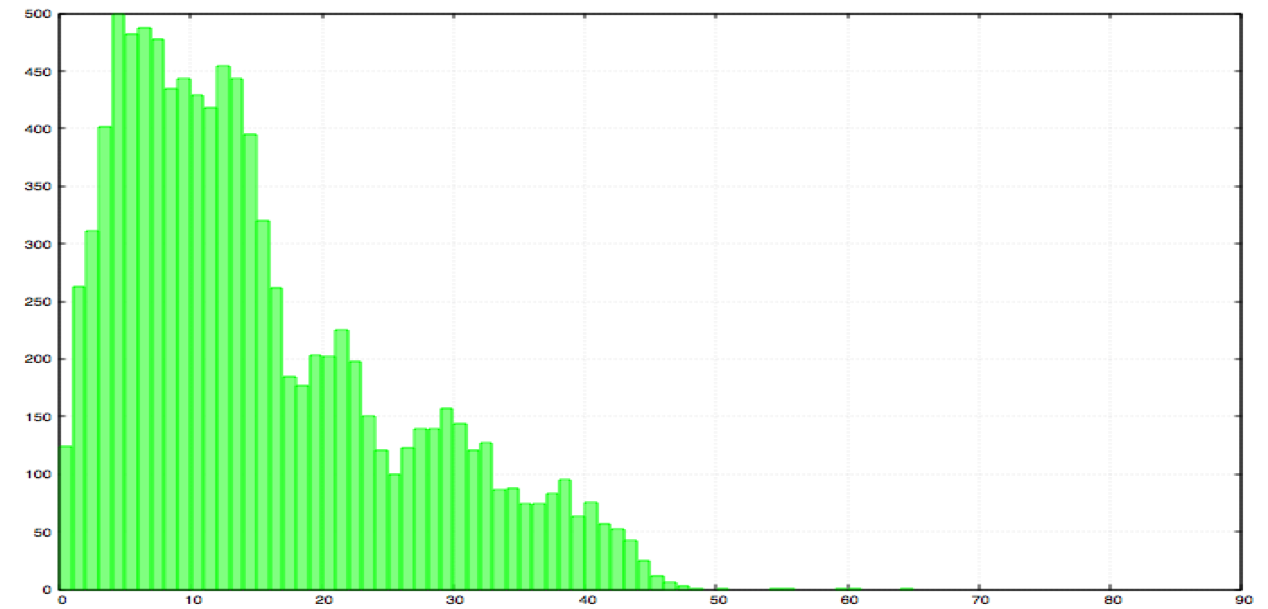}
	\caption{Distribution of different missing parts on the video dataset.}
	\label{fig:blockhists}
\end{center}
\end{figure}

\noindent
Missing parts (i.e., captioned area) in the video dataset have various sizes, since overlays size vary with the font size as well as length of subtitles. Figure~\ref{fig:blockhists} shows the distribution of missing parts ratio, highlighting the various difficulty of video samples, up to 50\% of frames to replace. Variations in number of pixels makes the proposed task challenging but realistic.

\noindent
We provide metadata with this dataset, consisting of every video sources, timestamped subtitles text, overlay size and background style for each video clip. Such metadata allows for analyzing systems performance according to overlays variations.

\subsection{Baselines}

The video track's main objective can be achieved by multiple auxiliary tasks, i.e.: detect regions that contain text, video segmentation, video/patch generation, etc. We implemented baselines from recent state of the art methods from the image inpainting domain. We adapted models from ~\cite{pathakCVPR16context, Yang_2017_CVPR, yeh2017semantic}, which  involve deep convolutional encoder-decoder architectures with reconstruction loss. We do not use temporal information, neither adversarial loss, giving room for lots of possible improvements during the challenge.

\begin{itemize}	
    \item {\bf Baseline 1} considers the whole frames as global context to infer missing parts, such as the complete spatial structure  to generate complete frames. We use a deep convolutional autoencoder, with a VGG-like encoder consisting of 3 blocks of 2 convolutional layers, batch normalization and relu non-linearities, with respectively 64, 128 and 256 filters of size 3x3 pixels. The decoder is composed of 4 blocks of convolutional, upsampling, batch normalization and relu activation. This model is heavy in terms of parameters and memory usage, since it assumes inputs and outputs of size 256x256x3.
	\item {\bf Baseline 2} processes frames by patches of 32x32 pixels, allowing to reduce the number of parameters. A patch auto-encoder (similar but lighter than first baseline) is jointly learned with a patch classifier that detects the presence of text. At inference, a frame is first split into non-overlapped patches. Then if the classifier informs about the presence of text, the patch is passed through the decoder. Patches classified as non-text are simply copied to the generated output frame. This approach is faster to train and efficient to remove text with no background overlays. However, since it ignores the context, it cannot perform well on patches containing only missing values.
\end{itemize} 

\subsection{Competition results}

Tables~\ref{validationTable2} and~\ref{testTable2} show the results of this track obtained in both validation and test phases of the Challenge, respectively. We report three evaluated metrics: MSE, peak signal-to-noise ratio (PSNR) and structural similarity index (SSIM). 
During the first phase, 35 different participants registered and downloaded the starting kit. Among them, 18 did at least one submission (not always valid), and 7 participants kept active. In total, we received 50 submissions, Table~\ref{validationTable2} and ~\ref{testTable2} list scores reached by active participants. \\

\begin{table}[]
\centering
\caption{Validation Results of track 2.}
\label{validationTable2}
\resizebox{\linewidth}{!}{\begin{tabular}{@{}llllll@{}}
\toprule
\textbf{\#} & \textbf{User} & \textbf{Rank} & \textbf{MSE} & \textbf{PSNR} & \textbf{SSIM} \\ \midrule
1           & \textbf{hcilab}    	& 3.3           & 0.0013 (1)   & 32.6001 (1)   & 0.0439 (8)    \\
2           & \textbf{arnavkj95}    & 3.6           & 0.0014 (2)   & 31.9629 (2)   & 0.0512 (7)    \\
3           & \textbf{dhkim}        & 4.0           & 0.0015 (3)   & 31.0613 (3)   & 0.0516 (6)    \\
4           & anubhap93  			& 4.3           & 0.0016 (6)   & 30.5311 (4)   & 0.0610 (5)    \\
5           & mcahny01     		    & 4.6           & 0.0018 (5)   & 29.9306 (6)   & 0.0751 (3)    \\
6           & ucs    			 	& 4.6           & 0.0021 (4)   & 28.6957 (7)   & 0.0935 (1)    \\
7           & \textbf{baseline2}	& 5.3           & 0.0023 (7)   & 30.0993 (5)   & 0.0621 (4)    \\
8           & yangchris11    		& 6.0           & 0.0045 (8)   & 27.0632 (8)   & 0.0876 (2)    \\
\bottomrule
\end{tabular}}
\end{table}

At the test phase, some participants merged their submissions, constituting a new team, while some did not submit any factsheets or codes as required to officially claim prizes. Table~\ref{testTable2} list every participants submission (in italic we show discarded participants).
\begin{table}[]
\centering
\caption{Test set Results of track 2.}
\label{testTable2}
\resizebox{\linewidth}{!}{\begin{tabular}{@{}llllll@{}}
\toprule
\textbf{\#} & \textbf{User} & \textbf{Rank} & \textbf{MSE} & \textbf{PSNR} & \textbf{SSIM} \\ \midrule
1           & \textbf{SanghyunWoo}    	& 2.6           & 0.0011 (1)   & 33.3527 (1)   & 0.0404 (6)    \\
2           & \textbf{hcilab}  		    & 3.0           & 0.0012 (3)   & 33.0228 (2)   & 0.0424 (4)    \\
3           & \textbf{\textit{ucs}}    	& 3.3           & 0.0011 (2)   & 33.0052 (3)   & 0.0410 (5)    \\
4           & \textbf{anubhap93}  		& 3.6           & 0.0012 (4)   & 32.0021 (5)   & 0.0499 (2)    \\
5           & arnavkj95     		    & 4.0           & 0.0012 (5)   & 32.1713 (4)   & 0.0482 (3)    \\
6           & \textbf{baseline2}		& 4.3           & 0.0022 (6)   & 30.1856 (6)   & 0.0613 (1)    \\
\bottomrule
\end{tabular}}
\end{table}

\subsection{Discussion}

Most of active participants did succeed in improving baseline methods. Next we briefly discuss on differences of proposed methods and their impact on inpainting performance.\\

\begin{itemize}	
\item{\textbf{SanghyunWoo}} used a U-Net based architecture combined with 3D and 2D gated convolutions to predict the middle frame of N given frames. Gated convolution allows for learning an attention mechanism in intermediate feature maps, while temporal dependence aims at taking advantage of the video structure. Training is performed by optimizing a loss that combines L1 reconstruction term, SSIM perceptual score, and Gradient L1 loss. \\
\item{\textbf{hcilab}} trained a U-Net model in a multi-stage procedure, where output of the previous stage is given as input of next stage, in a recurrent manner. The use of skip connections from U-Net in addition to an iterative scheme increase significantly the inpainting performance, compared to initial baseline2 performance. \\
\item{\textbf{anubhap93}} used an almost standard U-Net architecture with dilated convolutions. Their model was trained by combining MSE, SSIM and GAN losses. Dilated convolutions seem to significantly improve the generation quality. \\
\item{\textbf{arnavkj95}} used a standard U-Net architecture, trained with MSE loss. \\
\end{itemize}

We notice that only the winning team did take account of temporal context in video sequences to perform decaptioning. However, it is not clear whether they reached top performance due to this characteristic, since they are also the only team using gated convolutions. GANs were not intensively considered during this challenge, only one team combined Adversarial loss with MSE and SSIM losses, with a sensibly low impact, probably due to the difficulty and unstable nature of such training. The U-Net model has been widely considered as an effective encoder-decoder architecture. Most participants extended this model to improve inpainting performance. \\

\begin{figure*}[htbp]
\begin{center}
	\includegraphics[width=0.7\textwidth]{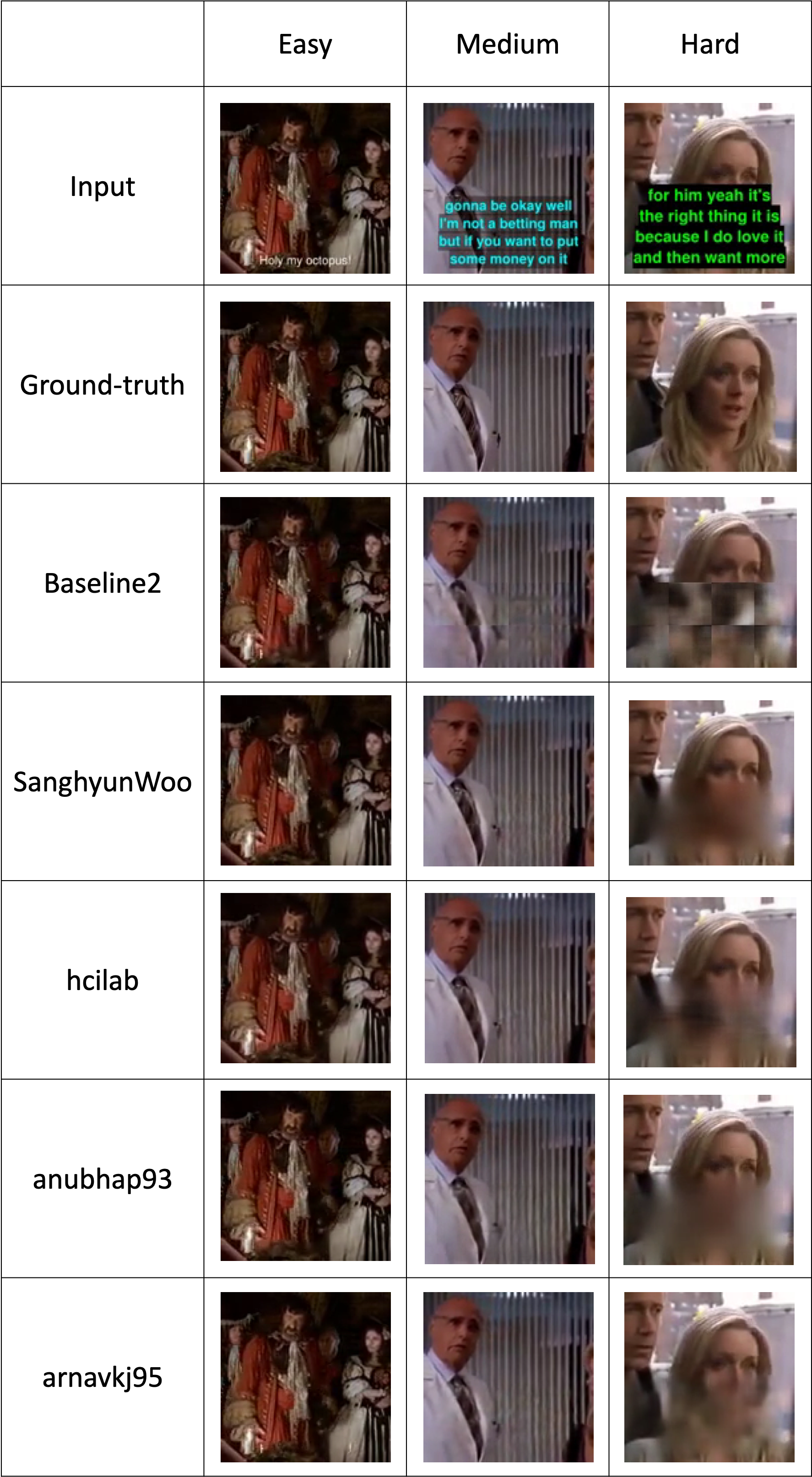}
	\caption{Qualitative comparison from the middle frame of three test samples.}
	\label{fig:track2_quali}
\end{center}
\end{figure*}

Figure~\ref{fig:track2_quali} shows a qualitative comparison of participants outputs for three different examples that belong to three levels of difficulty with respect to the size of missing part to inpaint: \textit{easy} samples mostly contain subtitles with small font, transparent background and few text; \textit{medium} samples have bigger font, semi-transparent background, more text ; \textit{difficult} samples contain a big part to replace (up to 50\% of frame) corresponding to subtitles with black-opaque background. We note that  all methods perform well for easy sample. From the medium sample, we remark small texture variations in the curtain reconstruction from participants outputs, where the baseline output fails to reconstruct well the right texture. The U-Net architecture seems here to bring a significant positive impact, since only our baseline method did not use any skip connections within the encoder-decoder architecture. Hard samples are extreme case of inpainting problem where a large part must be replaced. All the methods fail in getting acceptable reconstruction and generate blurry pixels. However, watching carefully, it appears that all the methods (except baseline) succeed in reconstructing the global structure of missing parts (face, hairs, neck, jacket). More carefully, we can observe that 'hcilab' seemed to generate details slightly better than others, especially considering the shape of face, where 'arnavkj95' missed more details than others. The blurry effect in this hard case might mostly come from the MSE loss used by most participants. In this case, GANs objective should help in generating sharper and more accurate regions.

\section{Fingerprint reconstruction and denoising} \label{sec:track3}

Biometrics play an increasingly important role in security. They ensure privacy and identity verification, as evidenced by the increasing prevalence of fingerprint sensors on mobile devices. Fingerprint retrieval keeps also being an important law enforcement tool used in forensics. However, much remains to be done to improve the accuracy of verification, both in terms of false negatives (in part due to poor image quality when fingers are wet or dirty) and false positives (due to the ease of forgery). This track involves reducing noise (denoising) and/or replacing missing parts (inpainting) due to various kinds of alterations or sensor failures in fingerprint images. This is viewed as a preprocessing step to ease verification carried out either by humans or existing third party software. To protect privacy and have full control over the experimental design, we synthesized a (very) large dataset of realistic artificial fingerprints for the purpose of training learning machines.

Thus, the objective of participants was to develop algorithms that can inpaint and denoise fingerprint images that contain artifacts like noise, scratches, etc. Such procedures can be applied to improve the performance of subsequent operations like fingerprint verification. Developed algorithms were evaluated based on reconstruction performance. That is, participants were required to reconstruct the degraded fingerprint images using their developed algorithms and submit the reconstructed fingerprint images. After the submission, the reconstructed  images were compared against the corresponding ground-truth fingerprint images in the pixel space to determine the quality of the reconstructions.

\subsection{Data}

We generated images of fingerprints by first degrading synthetic fingerprints with a distortion model (blur, brightness, contrast, elastic transformation, occlusion, scratch, resolution, rotation), then overlaying the fingerprints on top of various backgrounds. The resulting images were typical of what law enforcement agents have to deal with. For training participants got pairs of original and distorted images. Their goal was to recover original images from distorted image on a test dataset. 

Training set was constructed in the following two steps:  i) 75,600 $275\times400$ pixel ground-truth fingerprint images without any noise or scratches, but with random transformations (at most five pixels translation and +/-10 degrees rotation) were generated by using the software Anguli: Synthetic Fingerprint Generator. ii) 75,600 $275\times400$ pixel degraded fingerprint images were generated by applying random artifacts (blur, brightness, contrast, elastic transformation, occlusion, scratch, resolution, rotation) and backgrounds to the ground-truth fingerprint images. In total, it contains 151,200 fingerprint images (75,600 fingerprints, and two impressions - one ground-truth and one degraded - per fingerprint).

Validation and test sets were constructed similarly to the training set and were used to evaluate the reconstruction performance. In total, each set contains 16,800 fingerprint images (8,400 fingerprints and two impressions - one ground-truth and one degraded - per fingerprint). Since this set was used for the purpose of evaluating the reconstruction performance, only the degraded and not the ground-truth fingerprint images were provided to participants.

\subsection{Baselines}

A deep learning based baseline was used for reconstructing/enhancing the degraded fingerprint images with a standard deep convolutional neural network (CNN). To this end, a CNN was trained on the synthetic training set by minimizing the MSE loss function with Adam. The CNN comprises four convolutional layers, five residual blocks, two deconvolutional layers and another convolutional layer. Each of the five residual blocks comprises two convolutional layers. All of the layers except for the last layer are followed by batch normalization and rectified linear units. The last layer is followed by batch normalization and hyperbolic tangent units. The model is implemented in Chainer. 

\subsection{Competition results}

\begin{figure*}[h]
\begin{center}
	\includegraphics[width=1\textwidth]{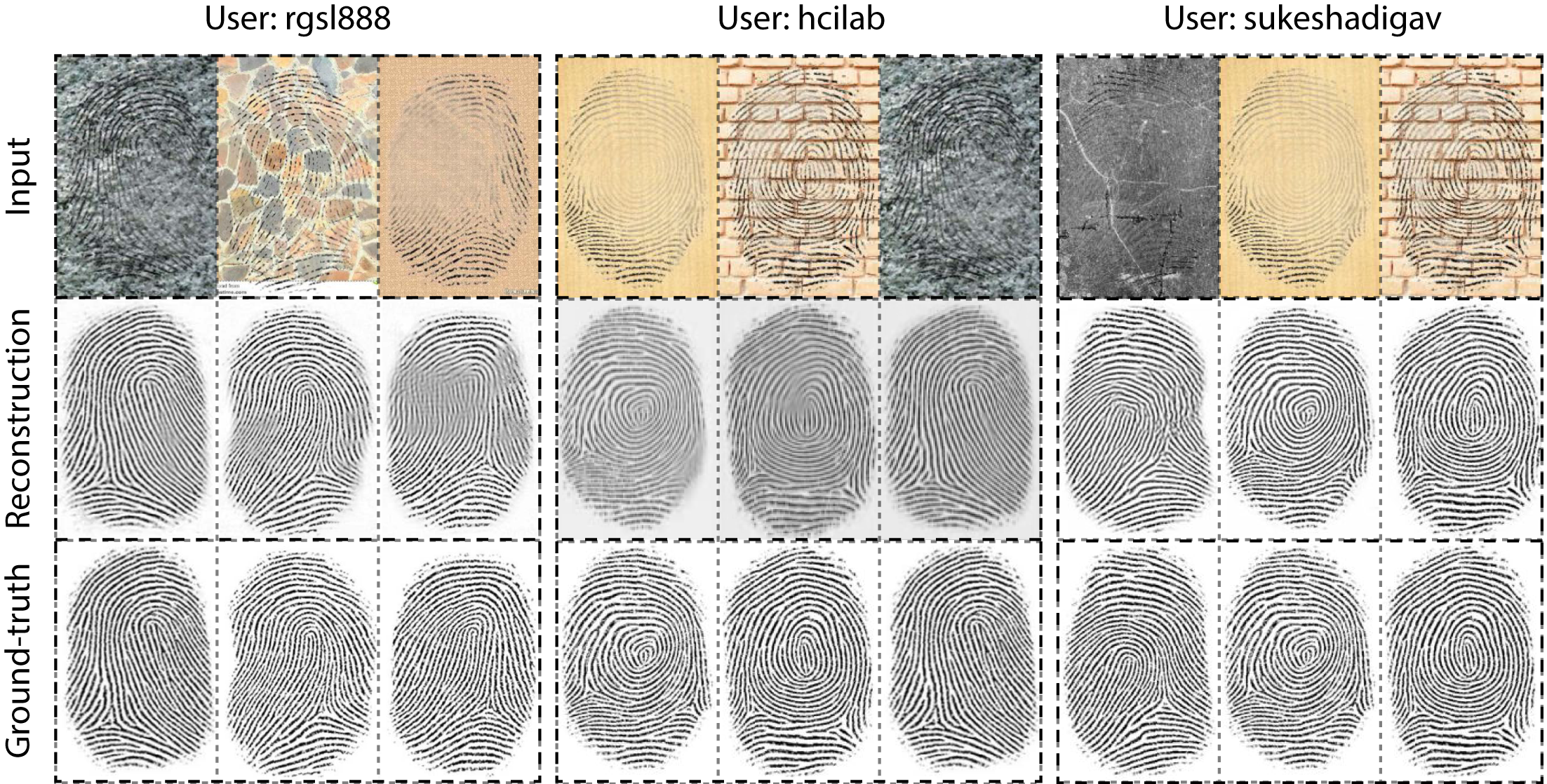}
	\caption{Sample results of three Track 3 submissions.}
	\label{fig:track_3_samples}
\end{center}
\end{figure*}

Similar to the other tracks, the results of this track were evaluated based on three different metrics. The metrics were mean squared error (MSE), peak signal-to-noise ratio (PSNR) and structural similarity index (SSIM). Table \ref{validationTable3} and \ref{testTable3} show the final results of the validation phase and the test phase. Figure \ref{fig:track_3_samples} shows qualitative results of a number of submissions. In total, there were 11 submissions in the validation phase and five submissions in the test phase. In the test phase, only four out of a potential 10 submissions outperformed the baseline, whereas in the validation phase, the baseline was outperformed by all four submissions. Interestingly, the only submissions in the test phase were those that were ranked above the baseline in the validation phase.

\begin{table}[]
\centering
\caption{Validation Results of track 3.}
\label{validationTable3}
\resizebox{\linewidth}{!}{\begin{tabular}{@{}llllll@{}}
\toprule
\textbf{\#} & \textbf{User} & \textbf{Rank} & \textbf{MSE} & \textbf{PSNR} & \textbf{SSIM} \\ \midrule
1           & \textbf{CVxTz}         & 1.0           & 0.0191 (1)   & 17.6640 (1)   & 0.8426 (1)    \\
2           & \textbf{rgsl888}       & 2.3           & 0.0239 (2)   & 16.8363 (2)   & 0.8069 (3)    \\
3           & \textbf{hcilab}        & 3.3           & 0.0241 (3)   & 16.6062 (3)   & 0.8031 (4)    \\
4           & sukeshadigav  & 4.0           & 0.0276 (6)   & 16.4162 (4)   & 0.8234 (2)    \\
5           & baseline      & 5.0           & 0.0252 (5)   & 16.4098 (5)   & 0.7954 (5)    \\
6           & finlouarn     & 5.3           & 0.0251 (4)   & 16.3992 (6)   & 0.7904 (6)    \\
7           & Xiaojing      & 7.0           & 0.0381 (7)   & 14.6347 (7)   & 0.6990 (7)    \\
8           & BriceRauby    & 8.0           & 0.0398 (8)   & 14.1740 (8)   & 0.6954 (8)    \\
9           & go            & 9.0           & 0.0414 (9)   & 14.1710 (9)   & 0.6709 (9)    \\
10          & yashkotadia   & 10.0          & 0.0564 (10)  & 12.7785 (10)  & 0.6417 (10)   \\
11          & yg            & 11.0          & 0.7282 (11)  & 1.3781 (11)   & 0.0001 (11)   \\ \bottomrule
\end{tabular}}
\end{table}

\begin{table}[]
\centering
\caption{Test results of track 3.}
\label{testTable3}
\resizebox{\linewidth}{!}{\begin{tabular}{@{}llllll@{}}
\toprule
\textbf{\#} & \textbf{User}         & \textbf{Rank} & \textbf{MSE} & \textbf{PSNR} & \textbf{SSIM} \\ \midrule
1           & \textbf{CVxTz}        & 1.0           & 0.0189 (1)   & 17.6968 (1)   & 0.8427 (1)    \\
2           & \textbf{rgsl888}      & 2.3           & 0.0231 (2)   & 16.9688 (2)   & 0.8093 (3)    \\
3           & \textbf{hcilab}       & 3.3           & 0.0238 (3)   & 16.6465 (3)   & 0.8033 (4)    \\
4           & \textbf{sukeshadigav} & 3.3           & 0.0268 (4)   & 16.5534 (4)   & 0.8261 (2)    \\
5           & baseline           & 5.0           & 0.0241 (5)   & 16.4160 (5)    & 0.8234 (5)    \\ \bottomrule
\end{tabular}}
\end{table}

\begin{table*}[]
\centering
\caption{Summary of the winning methods.}
\label{conclusionTable3}
\begin{tabular}{@{}lllll@{}}
\toprule
\textbf{\#} & \textbf{User} & \textbf{Model} & \textbf{Preprocessing} & \textbf{Training} \\ \midrule
\textbf{1} & \textbf{CVxTz} & U-Net & \begin{tabular}[c]{@{}l@{}}Normalization,\\ rescaling,\\ resizing\end{tabular} & \begin{tabular}[c]{@{}l@{}}Vanilla backprop with MSE,\\ data augmentation\\ (affine, contrast, saturation)\end{tabular} \\
\textbf{2} & \textbf{rgsl888} & \begin{tabular}[c]{@{}l@{}}U-Net,\\ dilated conv\end{tabular} & - & Vanilla backprop with MSE \\
\textbf{3} & \textbf{hcilab} & Baseline & - & Iterative backprop with MSE \\
\textbf{4} & \textbf{sukeshadigav} & M-Net & \begin{tabular}[c]{@{}l@{}}Normalization,\\ padding,\\ resizing\end{tabular} & \begin{tabular}[c]{@{}l@{}}Vanilla backprop with MSE,\\ SSIM\end{tabular} \\
5 & baseline & Residual encoder & Resizing & Vanilla backprop with MSE \\ \bottomrule
\end{tabular}
\end{table*}

\subsection{Discussion}

There were a number of similarities and differences between the winning methods and the baseline. A summary thereof is presented in Table \ref{conclusionTable3}.

Three out of four of the winning methods used a U-net like architecture and the remaining method used the same architecture as the baseline. Given that the most striking difference between the baseline and such U-net like architectures are skip connections and bottleneck layers, these results demonstrate the importance of these architectural components in the task of inpainting and denoising for fingerprint verification.

Next to the architectural differences, there were other important factors that differentiated the higher ranking methods from the rest. In particular, it seems that the extensive use of data augmentation by the first placed team and the use of dilated convolutions by the second placed team were the main factors that separated them from the rest of the teams.

Another interesting observation was that almost all teams exclusively used the MSE loss function, which is notorious for causing artefacts in image generation tasks, suggesting that the noise ceiling on this dataset was not reached as the methods were held back by suboptimal loss functions.

Taken together, the results are expected to be improved with more realistic forward models, adversarial/feature loss functions and perhaps variational methods. It would also be interesting to see how the current methods or such improved methods compare with physical models and transfer to real data.

\section{Conclusion and future directions of research} \label{sec:conclusion}
%Hugo
We have described the design and results of a novel challenge focusing on inpainting in images and video. Three tracks were proposed, each associated to a realistic scenario. For each track we prepared new datasets and baselines, exposing the complexity of the tasks and showing the feasibility of solving them. Results obtained by participants were presented and analyzed. In all tracks, participants succeeded at improving the performance of baseline methods. Also, it is impressive the qualitative performance of proposed solutions. 
 
We can outline the following conclusions and corresponding future research directions as derived from the challenge:
\begin{itemize}
\item \textbf{Methodological similarity among solutions.} Results obtained by participants together with the undeniable similarity among solutions, all of them based on deep learning, confirm the establishment of specific architectures of deep learning (e.g., U-NET like architectures) for visual inpainting. In fact, the difference among solutions and their performance in some cases was only due to hyperparameter or subtle changes in the overall architecture of the models. This finding seems to suggest that a promising line of future research is that of the \emph{automatic construction of deep learning models} (AutoDL)~\cite{AutoDL}. That is, methods that automatically can determine the architecture and hyperparameters of a model given a specific task. This is in fact a growing area not only in terms of academic research but of industrial importance. 

\item \textbf{Feasibility of the approached tasks and potential impact.} The quantitative, but mainly the qualitative performance of solutions evidences the feasibility of the approached task: for the three proposed tracks, top ranked solutions represent satisfactory solutions that could be put in practice in short time. In this sense, the challenge was successful in providing academically challenging problems, but also in organizing tracks around problems of practical relevance. What is more, solutions are based on what can be considered domain public knowledge. This will surely motivate other researchers and practitioners to approach similar tasks.

\item \textbf{Quality and quantity on newly released datasets.} As previously mentioned, a major contribution of the organization of the inpainting challenge lies in the resources being generated. We provided novel datasets, associated to realistic scenarios with a large potential of application. The released datasets are now public and comprise the largest resources to date in their corresponding application. We foresee these datasets will be decisive in the progress of the field in the forthcoming years. Likewise, we think that a promising line of research has to do with the generation of resources for visual inpainting, including novel ways of labeling huge amounts of data. 

\item \textbf{Evaluation metric / loss function.} Although we have used SSIM as structural similarity metric, the qualitative results show such a metric is not sufficient to evaluate realistic inpainting and perceptions. Training a network with SSIM or VGG-based perceptual loss can not guarantee a realistic inpainting to the unseen test data. This can be an interesting and important research topic in the future.

\end{itemize}

\section*{Acknowledgements}

The sponsors of ChaLearn Looking at People inpainting and denoising events are Google, ChaLearn, Amazon, and Disney Research. This work has been partially supported by the Spanish project TIN2016-74946-P (MINECO/FEDER, UE) and CERCA Programme / Generalitat de Catalunya. This work was also partially funded by the French national research agency (grant number ANR16-CE23-0006). We gratefully acknowledge the support of NVIDIA Corporation with the donation of the GPU used for this research. This work is partially supported by ICREA under the ICREA Academia programme. We thank all challenge participants for their excellent contributions.

%\nocite{*}

\bibliography{main}

\begin{thebibliography}{10}

\bibitem{statista}
Hours of video uploaded to youtube every minute as of july 2015.
\newblock
  \url{https://www.statista.com/statistics/259477/hours-of-video-uploaded-to-youtube-every-minute/},
  2019.

\bibitem{andriluka14cvpr}
Mykhaylo Andriluka, Leonid Pishchulin, Peter Gehler, and Bernt Schiele.
\newblock 2d human pose estimation: New benchmark and state of the art
  analysis.
\newblock In {\em IEEE Conference on Computer Vision and Pattern Recognition
  (CVPR)}, June 2014.

\bibitem{charles2013domain}
James Charles, Tomas Pfister, Derek~R Magee, David~C Hogg, and Andrew
  Zisserman.
\newblock Domain adaptation for upper body pose tracking ian signed tv
  broadcasts.
\newblock In {\em BMVC}, 2013.

\bibitem{dong2016image}
Chao Dong, Chen~Change Loy, Kaiming He, and Xiaoou Tang.
\newblock Image super-resolution using deep convolutional networks.
\newblock {\em IEEE transactions on pattern analysis and machine intelligence},
  38(2):295--307, 2016.

\bibitem{eichner2012human}
Marcin Eichner and Vittorio Ferrari.
\newblock Human pose co-estimation and applications.
\newblock {\em IEEE transactions on pattern analysis and machine intelligence},
  34(11):2282--2288, 2012.

\bibitem{he2016deep}
Kaiming He, Xiangyu Zhang, Shaoqing Ren, and Jian Sun.
\newblock Deep residual learning for image recognition.
\newblock In {\em Proceedings of the IEEE conference on computer vision and
  pattern recognition}, pages 770--778, 2016.

\bibitem{jain2009natural}
Viren Jain and Sebastian Seung.
\newblock Natural image denoising with convolutional networks.
\newblock In {\em Advances in Neural Information Processing Systems}, pages
  769--776, 2009.

\bibitem{slomo-nvidia}
Huaizu Jiang, Deqing Sun, Varun Jampani, Ming{-}Hsuan Yang, Erik~G.
  Learned{-}Miller, and Jan Kautz.
\newblock Super slomo: High quality estimation of multiple intermediate frames
  for video interpolation.
\newblock In {\em 2018 {IEEE} Conference on Computer Vision and Pattern
  Recognition, {CVPR} 2018, Salt Lake City, UT, USA, June 18-22, 2018}, pages
  9000--9008, 2018.

\bibitem{Johnson10}
Sam Johnson and Mark Everingham.
\newblock Clustered pose and nonlinear appearance models for human pose
  estimation.
\newblock In {\em Proceedings of the British Machine Vision Conference}, 2010.
\newblock doi:10.5244/C.24.12.

\bibitem{AutoDL}
Zhengying Liu, Olivier Bousquet, Andr\'e Elisseeff, Sergio Escalera, Isabelle
  Guyon, Julio~Jacques Jr., Adrien Pavao, Danny Silver, Lisheng Sun-Hosoya,
  Sebastien Treguer, Wei-Wei Tu, Jingsong Wang, and Quanming Yao.
\newblock Autodl challenge design and beta tests: towards automatic deep
  learning.
\newblock In {\em Submitted to NIPS Workshop on Meta-Learning}, 2018.

\bibitem{mao2016image}
Xiao-Jiao Mao, Chunhua Shen, and Yu-Bin Yang.
\newblock Image restoration using convolutional auto-encoders with symmetric
  skip connections.
\newblock {\em arXiv preprint arXiv:1606.08921}, 2016.

\bibitem{newell2016stacked}
Alejandro Newell, Kaiyu Yang, and Jia Deng.
\newblock Stacked hourglass networks for human pose estimation.
\newblock In {\em European Conference on Computer Vision}, pages 483--499.
  Springer, 2016.

\bibitem{newson2014video}
Alasdair Newson, Andr{\'e}s Almansa, Matthieu Fradet, Yann Gousseau, and
  Patrick P{\'e}rez.
\newblock Video inpainting of complex scenes.
\newblock {\em SIAM Journal on Imaging Sciences}, 7(4):1993--2019, 2014.

\bibitem{pathakCVPR16context}
Deepak Pathak, Philipp Kr\"ahenb\"uhl, Jeff Donahue, Trevor Darrell, and Alexei
  Efros.
\newblock Context encoders: Feature learning by inpainting.
\newblock In {\em Computer Vision and Pattern Recognition ({CVPR})}, 2016.

\bibitem{modec13}
Benjamin Sapp and Ben Taskar.
\newblock Modec: Multimodal decomposable models for human pose estimation.
\newblock In {\em In Proc. CVPR}, 2013.

\bibitem{wang2004image}
Zhou Wang, Alan~C Bovik, Hamid~R Sheikh, and Eero~P Simoncelli.
\newblock Image quality assessment: from error visibility to structural
  similarity.
\newblock {\em IEEE transactions on image processing}, 13(4):600--612, 2004.

\bibitem{xie2012image}
Junyuan Xie, Linli Xu, and Enhong Chen.
\newblock Image denoising and inpainting with deep neural networks.
\newblock In {\em Advances in neural information processing systems}, pages
  341--349, 2012.

\bibitem{xu2014deep}
Li~Xu, Jimmy~SJ Ren, Ce~Liu, and Jiaya Jia.
\newblock Deep convolutional neural network for image deconvolution.
\newblock In {\em Advances in Neural Information Processing Systems}, pages
  1790--1798, 2014.

\bibitem{Yang_2017_CVPR}
Chao Yang, Xin Lu, Zhe Lin, Eli Shechtman, Oliver Wang, and Hao Li.
\newblock High-resolution image inpainting using multi-scale neural patch
  synthesis.
\newblock In {\em The IEEE Conference on Computer Vision and Pattern
  Recognition (CVPR)}, July 2017.

\bibitem{yeh2017semantic}
Raymond~A. Yeh$^\ast$, Chen Chen$^\ast$, Teck~Yian Lim, Schwing~Alexander G.,
  Mark Hasegawa-Johnson, and Minh~N. Do.
\newblock Semantic image inpainting with deep generative models.
\newblock In {\em Proceedings of the IEEE Conference on Computer Vision and
  Pattern Recognition}, 2017.
\newblock $^\ast$ equal contribution.

\end{thebibliography}

\end{document}